\documentclass{article} 
\usepackage{nips10submit_e,times,graphicx}

\title{Efficient Learning of Sparse Invariant Representations}

\author{
Karol Gregor and Yann LeCun\\
Courant Institute, New York University
New York, NY, 10003, USA\\
\texttt{kgregor@cs.nyu.edu, yann@cs.nyu.edu} \\
}

\nipsfinalcopy 

\begin{document}

\maketitle

\begin{abstract}
We propose a simple and efficient algorithm for learning sparse
invariant representations from unlabeled data with fast
inference. When trained on short movies sequences, the learned
features are selective to a range of orientations and spatial
frequencies, but robust to a wide range of positions, similar to
complex cells in the primary visual cortex. We give a hierarchical
version of the algorithm, and give guarantees of fast convergence
under certain conditions.\footnote[1]{This paper was completed in June 2010, and submitted (unsuccessfully) to NIPS 2010.}
\end{abstract}

\section{Introduction}

Learning representations that are invariant to irrelevant
transformations of the input is an important step towards building
recognition systems automatically. Invariance is a key property of
some cells in the mammalian visual cortex. Cells in high-level areas
of the visual cortex respond to objects categories, and are invariant
to a wide range of variations on the object (pose, illumination,
confirmation, instance, etc). The simplest known example of invariant
representations in visual cortex are the complex cells of V1 that
respond to edges of a given orientation but are activated by a wide
range of positions of the edge. Many artificial object recognition
systems have built-in invariances, such as the translational
invariance of convolutional network~\cite{lecun-89e}, or SIFT
descriptors \cite{lowe2004distinctive}. An important question is how
can useful invariant representations of the visual world be learned
from unlabeled samples.

In this paper we introduce an algorithm for learning features that are
invariant (or robust) to common image transformations that typically
occur between successive frames of a video or statistically within a single frame. While the method is quite
simple, it is also computationally efficient, and possesses provable
bounds on the speed of inference.  The first component of the model is
a layer of sparse coding. Sparse coding~ \cite{olshausen1996emergence}
constructs a dictionary matrix $W$ so that input vectors can be
represented by a linear combination of a small number of columns of
the dictionary matrix. Inference of the feature vector $z$
representing an input vector $x$ is performed by finding the
$z$ that minimizes the following energy function
\begin{eqnarray}
\label{eq_sparse_coding}
E_1(W,x,z) &=& \frac{1}{2} \|x-Wz\|^2 + \alpha |z|_{L1}
\end{eqnarray}
where $\alpha$ is a positive constant.  The dictionary matrix $W$ is
learned by minimizing $\min_z E(W,x^k,z)$ averaged over a set of
training samples $x^k \;\;\; k=1\ldots K$, while constraining the
columns of $W$ to have norm 1.  

The first idea of the proposed method
is to accumulate sparse feature vectors representing successive frames
in a video, or versions of an image that are distorted by
transformations that do not affect the nature of its content.
\begin{eqnarray}
\label{eq_temporel_sparse_coding}
z^* &=& \sum_t \left| {\rm argmin}_z \frac{1}{2} \|x_t-Wz\|^2 + \alpha |z|_{L1} \right|
\end{eqnarray}
where the sum runs over the distorted images $x_t$.  
The second idea
is to connect a second sparse coding layer on top of the first one
that will capture dependencies between components of the accumulated
sparse code vector. This second layer models vector $z^*$ using
an {\em invariant code} $u$, which is the minimum of the following
energy function
\begin{eqnarray}
\label{eq_EI_gh}
E_2(A,z^*,u) &=& \alpha \sum_i z^*_i e^{(-Au)_i} + \beta |u| \\
\end{eqnarray}
where $|u|$ denotes the $L1$ norm of $u$, $A$ is a matrix, and $\beta$
is a positive constant controlling the sparsity of $u$.  Unlike with
traditional sparse coding, in this method the dictionary matrix
interacts {\em multiplicatively} with the input $z^*$. As in
traditional sparse coding, the matrix $A$ is trained by gradient
descent to minimize the average energy for the optimal $u$ over a
training set of vectors $z^*$ obtained as stated above. The columns of
$A$ are constrained to be normalized to 1. Essentially, the matrix $A$
will connect a component of $u$ to a set of components of $z^*$
if these components of $z^*$ co-occur frequently. When a component of
$u$ turns on, it has the effect of lowering the coefficients of the
components of $|z^*_i|$ to which it is strongly connected through the
$A$ matrix. To put it another way, if a set of components of $z^*$
often turn on together, the matrix $A$ will connect them to a
component of $u$. Turning on this component of $u$ will lower the
overall energy (if $\beta$ is small enough) because the whole set of
components of $z^*$ will see their coefficient being lowered (the
exponential terms). Hence, each unit of $u$ will connect units
of that often turn on together within a sequence of images. These
units will typically represent distorted version of a feature.

The energies (\ref{eq_sparse_coding}) and (\ref{eq_fullE}) can be naturally combined into a single combined model of $z$ and $u$ as explained in section \ref{sec_model}. There the second layer $u$ is essentially modulating sparsity of the first layer $z$. Single model of the image is more natural. For the invariance properties we didn't find much qualitative difference and since the former has provable inference bounds we presented the results for separate training. However the a two layer model should capture the statistics of an image. To demonstrate this we compared the in-paining capability of one and two layer models and found that two layer model does better job. For these experiments, the combined two layer model is necessary. We also found that despite the assumptions of the fast inference are not satisfied for the two layer model, empirically the inference is fast in this case as well.

\subsection{Prior work on Invariant Feature Learning}
The first way to implement invariance is to take a known invariance,
such as translational invariance in images, in put it directly into
the architecture. This has been highly successful in convolutional
neural networks \cite{lecun-89e} and SIFT descriptors
\cite{lowe2004distinctive} and its derivatives. The major drawback of
this approach is that it works for known invariances, but not unknown
invariances such as invariance to instance of an object. A system that
would discover invariance on its own would be desired.

Second type of invariance implementation is considered in the
framework of sparse coding or independent component analysis. The idea
is to change a cost function on hidden units in a way that would
prefer co-occurring units to be close together in some space
\cite{hyvarinen2001two,koray-cvpr-09}. This is achieved by pooling
units close in space together. This groups different inputs together
producing a form of invariance. The drawback of this approach is that
it requires some sort of imbedding in space and that the filters have
to arrange themselves.

In the third approach, rather then forcing units to arrange
themselves, we let them learn whatever representations they want to
learn and instead figure out which to pool together. In
\cite{karklin2003learning,karklin2008emergence}, this was achieved by
modulating covariance of the simple units with complex units.

The fourth approach to invariance uses the following idea: If the
inputs follow one another in time they are likely a consequence of the
same cause. We would like to discover that cause and therefore look
for representations that are common for all frames. This was achieved
in several ways. In slow feature analysis
\cite{wiskott2002slow,berkes2005slow,bergstra22slow} one forces the
representation to change slowly. In temporal product network
\cite{tpn} one breaks the input into two representations - one that is
common to all frames and one that is complementary. In
\cite{NIPS2008_0200} the idea is similar but in addition the
complementary representation specifies movement. In the simplest
instance of hierarchical temporal memory \cite{george2009towards} one
forms groups based on transition matrix between states. The
\cite{berkes2009structured} is a structured model of video.

A lot of the approaches for learning invariance are inspired by the
fact that the neo-cortex learns to create invariant
representations. Consequently these approaches are not focused on
creating efficient algorithms. In this paper, we given an efficient
learning algorithm that falls into the framework of third and fourth
approaches. The basic idea is to modulate the sparsity of sparse coding
units using higher level units that are also sparse. The fourth
approach is implemented by using the same higher level representation
for several consecutive time frames. In the form our model is similar
to that of \cite{karklin2003learning,karklin2008emergence} but a
little simpler. In a sense comparing our model to
\cite{karklin2003learning,karklin2008emergence} is similar to
comparing sparse coding to independent component analysis. Independent
component analysis is a probabilistic model, whereas sparse coding
attempts to reconstruct input in terms of few active hidden units. The
advantage of sparse coding is that it is simpler and easier to
optimize. There exist several very efficient inference and learning
algorithms
\cite{FISTA,li2009coordinate,hale2008fixed,lee-nips-06,mairal-icml-09}
and sparse coding has been applied to a large number problems. It is
this simplicity that allows efficient training of our model. The
inference algorithm is closely derived from the fast iterative
shrinkage-thresholding algorithm (FISTA) \cite{FISTA} and has a
convergence rate of $1/k^2$ where $k$ is the number of iterations.

\section{The Model}
\label{sec_model}

The model described above comprises two separately trained modules,
whose inference is performed separately. However, one can devise
a unified model with a single energy function that is conceptually
simpler:
\begin{eqnarray}
\label{eq_fullE}
E(W,A,x,z,u) &=& \frac{1}{2} \sum_t \|x_t-Wz_t\|^2 +
  \sum_i \sum_t \alpha |z_{ti}| g(u)_i + h(u)\\
g(u)_i &=& (1+e^{-(Au)_i})/2,\hspace{2mm} h(u)=\beta |u|
\nonumber
\end{eqnarray}
Given a set of inputs $x^{\gamma}$, the goal of training is to
minimize $\sum_{\gamma} E(W,A,x^{\gamma},z^{\gamma},u^{\gamma})$. We
do this by choosing one input $x$ at a time, minimizing
(\ref{eq_fullE}) over $z$ and $u$ with $W$ and $A$ fixed, then fixing
the resulting $z_{\min}$,$u_{\min}$ and taking step in a negative
gradient direction of $W$, $A$ (stochastic gradient descent). An
algorithm for finding $z_{\min}$,$u_{\min}$ is given in section
\ref{sec_theoretical}. It consists of taking step in $z$ and $u$
separately, each of which lowers the energy.

Note: The $g$ functions in (\ref{eq_fullE}) is different from that of
the simple (split) model. The reason is that, in our experiments,
either $u$ units lower the sparsity of $z$ too much, not resulting in
a sparse $z$ code or the units $u$ do not turn on at all.

\subsection{A Toy Example}
We now describe a toy example that illustrates the main idea of the
model~\cite{hinton-discuss}. The input, with $n_t=1$, is an image
patch consisting of a subset of the set of parallel lines of four
different orientations and ten different positions per
orientation. However for any given input, only lines with the same
orientation can be present, Figure \ref{fig_lines}a (different
orientations have equal probability and for a given orientation a line
of this orientation is present with probability 0.2 independently of
others). This is a toy example of a texture. Training sparse coding on
this input results in filters similar to one in Figure
\ref{fig_lines}b. We see that a given simple unit responds to one
particular line. The noisy filters correspond to simple units that are
inactive - this happens because there are only 40 discrete inputs. In
realistic data such as natural images, we have a continuum and
typically all units are used.

Clearly, sparse coding cannot capture all the statistics present in
the data. The simple units are not independent. We would like to learn
that that units corresponding to lines of a given orientation usually
turn on simultaneously. We trained (\ref{eq_fullE}) on this data
resulting in the filters in the Figure \ref{fig_lines}b,c. The filters
of the simple units of this full model are similar to those obtained
by training just the sparse coding. The invariant units pool together
simple units with filters corresponding to lines of the same
orientation. This makes the invariant units invariant to the pattern
of lines and dependent only on the orientation. Only four invariant
units were active corresponding to the four groups. As in sparse
coding, on a realistic data such as natural images, all invariant
units become active and distribute themselves with overlapping filters
as we will se below.

\begin{figure}[h]
\begin{center}
\centerline{\includegraphics[width=\columnwidth]{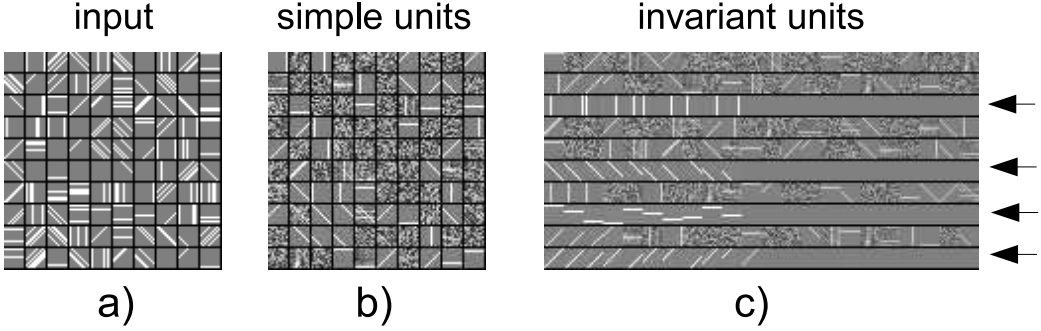}}
\end{center}
\caption{Toy example trained using model (\ref{eq_fullE}). a) Randomly
  selected input image patches. The input patches are generated as
  follows. Pick one of the four orientations at random. Consider all
  lines of this orientation. Put any such line into the image
  independently with probability 0.2. b) Learned sparse coding
  filters. A given active unit responds to a particular line. c)
  Learned filters of the invariant units. Each row corresponds to an
  invariant unit. The sparse coding filters are ordered according to
  the strength of their connection to the invariant unit. There are
  only four active units (arrows) and each responds to a given
  orientation, invariant to which lines of a given orientation are
  present.}
\label{fig_lines}
\end{figure}

Let us now discuss the motivation behind introducing a sequence of
inputs ($n_t>1$) in (\ref{eq_fullE}). Inputs that follow one another
in time are usually a consequence of the same cause. We would like to
discover that cause. This cause is something that is present at all
frames and therefore we are looking for a single representation $u$ in
(\ref{eq_fullE}) that is common to all the frames.

Another interesting point about the model (\ref{eq_fullE}) is a that
nonzero $u$ lowers the sparsity coefficient of units of $z$ that
belong to a group making them more likely to become activated. This
means that the model can utilize higher level information (which group
is present) to modulate the activity of the lower layer. This is a
desirable property for multi-layer systems because different parts of
the system should propagate their belief to other parts. In our
invariance experiments the results for the unified model were very similar to the results of the simple (split) model. Below we show the results of this
simple model because it is simple and because we provably know an
efficient inference algorithm. However in the section \ref{sec_theoretical} we will revisit the full system, generalize it to an $n$-layer system, give an
algorithm for training it, and prove that under some assumptions of
convexity, the algorithm again has a provably efficient inference. In the final section we use the full system for in-paining and show that it generalizes better then a single layer system.

\section{Efficient Inference for the Simplified Model}
\label{sec_simp_model}

Here we discuss how to find $u$ efficiently and give the numerical
results of the paper. The results for the full model (\ref{eq_fullE})
were similar.

\subsection{FISTA training algorithm}

The advantage of (\ref{eq_EI_gh}) compared to (\ref{eq_fullE}) is that
the fast iterative shrinkage-thresholding algorithm (FISTA)
\cite{FISTA} applies to it directly. FISTA applies to problems of the
form $E(u)=f(u)+\hat{g}(u)$ where:
\begin{itemize}
\item $f$ is continuously differentiable, convex and Lipschitz, that
  is $\| \nabla f(u_1) - \nabla f(u_2)\| \leq L(f)\|u_1-u_2\|$. The
  $L(f)>0$ is the Lipschitz constant of $\nabla f$.
\item $\hat{g}$ is continuous, convex and possibly non-smooth
\end{itemize}
The problem is assumed to have solution $E^{*}=E(u^{*})$. In our case
$f(u)=\alpha |z_t| g(u)$ and $\hat{g}(u)=h(u)$ which satisfies these
assumptions ($A$ is initialized with nonnegative entries which stay
nonnegative during the algorithm without a need to force it). This
solution converges with bound $E(u_k)-E(u^{*}) \leq 2 \alpha L(f)
\|u_0-u^{*}\|^2/(k+1)^2$ where $u_k$ is the value of $u$ at the
$k$-$th$ iteration and $\alpha$ is a constant. The cost of each
iteration is $O(mn)$ where $n$ is the input size and $m$ is the output
size. More precisely the cost is one matrix multiplications by $A$ and
by $A^t$ plus $O(m+n)$ cost. We used the back-tracking version of the
algorithm to find $L$ which contains a fixed number of $O(mn)$
operations (independent of desired error). It is a standard knowledge
and easy to see that the algorithm applies to the sparse coding
(\ref{eq_sparse_coding}) as well.

\subsection{Results}

The input to the network was prepared as follows. We converted all the
images of the Berkeley data-set into gray-scale images. We locally
removed the mean for each pixel by subtracting a Gaussian-weighted
average of the nearby pixels. The width of the Gaussian was $9$
pixels. Then, we locally normalized the contrast by dividing each
pixel by Gaussian-weighted standard deviation of the nearby pixels
(with a small cutoff to prevent blow-ups). The width of the Gaussian
was also $9$ pixels. Then, we picked a $20 \times 20$ window in the
image and, for a randomly chosen direction and magnitude, we moved it
for $n_t=3$ frames and extracted them. The magnitude of the
displacement was random in the range of $1-2$ pixels. A very large
collection of such triplets of frames was extracted. We trained the
sparse coding algorithm with $400$ code units in $z$ on each
individual frame (not on the $n_t$ concatenated frames). After
training we found the sparse codes for each frame. There were $100$
units in the layer of invariant units $u$. For larger a system with
$1000$ simple units and $400$ invariant units, see the supplementary
material.

\begin{figure}[h]
\begin{center}
\centerline{\includegraphics[width=\columnwidth]{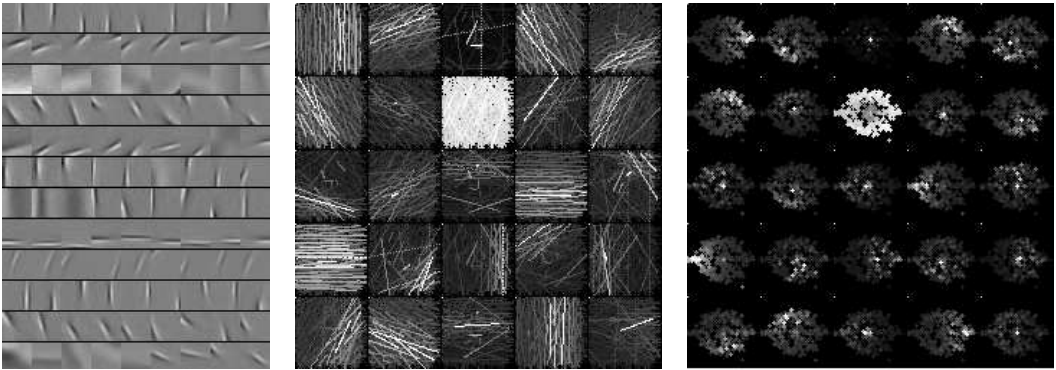}}
\end{center}
\caption{Connectivity between invariant units and simple units. {\bf
    Left:} Each row correspond to an invariant unit. For a given
  invariant unit, the filters of the simple units were ordered
  according to the size of the weight to the invariant unit. The
  strongest 9 are plotted. {\bf Middle:} Each square correspond to an
  invariant unit. 25 out of 100 invariant units were selected at
  random. The sparse coding filters were fitted with gabor
  functions. The center position of each line is the center of the
  Gabor filter. The orientation is the orientation of the sine-wave of
  the Gabor filter. The brightness is proportional to the strength of
  connection between invariant and simple units. For each invariant
  unit the weights were normalized for drawing so that the strongest
  connection is white and zero connections are black. {\bf Right:}
  Each square (circle) corresponds to an invariant unit. Each dot
  correspond to a simple unit. The distance from the center is the
  frequency of the Gabor fit. The angle is twice the orientation of
  the Gabor fit (twice because angles related by $\pi$ are
  equivalent). We see that invariant units typically learn to group
  together units with similar orientation and frequency. There are few
  other types of filters as well. The units in the middle and right
  panel correspond to each other and correspond to the units in the
  left panel reading panels left to right and then down. See the
  supplementary material for all the filters the system: $20 \times
  20$ input patches, $1000$ simple units, $400$ invariant units.}
\label{fig_rfu}
\end{figure}

The results are shown in the Figure \ref{fig_rfu}, see caption for
description. We see that many invariant cells learn to group together
filters of similar orientation and frequency but at several positions
and thus learn invariance with respect to translations. However there
are other types of filters as well. Remember that the algorithm learns
statistical co-occurrence between features, whether in time or in
space.

\begin{figure}[h]
\begin{center}
\centerline{\includegraphics[width=\columnwidth]{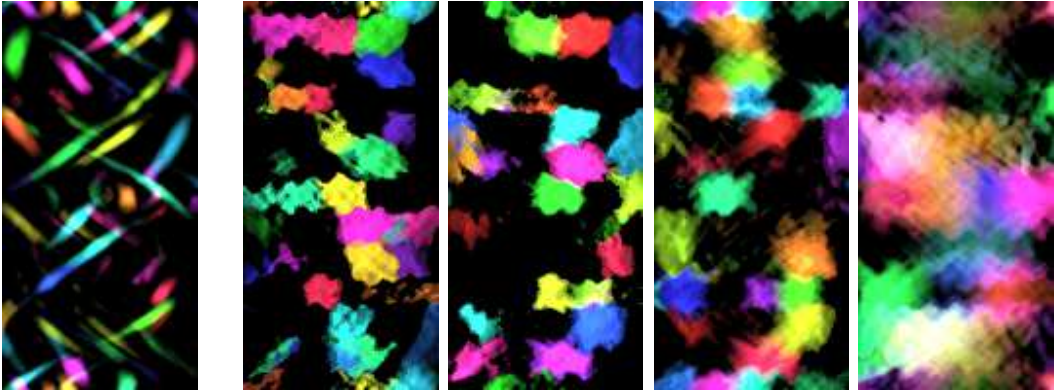}}
\end{center}
\caption{Responses of simple units and invariant units to the input
  edge from equation (\ref{eq_edge}). Left panel are responses of
  simple units trained with sparsity $\alpha=0.5$ in
  (\ref{eq_sparse_coding}). The right four panels are responses of invariant
  units trained with sparsities $\beta=0.5, 0.3, 0.2, 0.1$ in
  (3) on the values of simple units. The x-axis of each
  panel is the distance of the edge from the center of the image - the
  $b$ in (\ref{eq_edge}). The y-axis is the orientation of each edge -
  the $\theta$ in (\ref{eq_edge}). 30 cells were chosen at random in
  each panel. Different colors correspond to different cells. The
  color intensity is proportional to the response of the unit. We see
  that sparse coding inputs respond to a small range of frequencies
  and positions (the elongated shape is due to the fact that an edge
  of orientation somewhat different from the edge detector orientation
  sweeps the detector at different positions $b$). On the other hand
  invariant cells respond to edges of at similar range of frequencies
  but larger range of positions. At high sparsities the response
  boundaries are sharp and response regions don't overlap. As we lower
  the sparsity the boundaries become more blurry and regions start to
  overlap. $k=1$ was used in (\ref{eq_edge}). Other frequencies
  produced similar effect.}
\label{fig_responses}
\end{figure}

The values of the weights give us important information about the
properties of the system. However ultimately we are interested in how
the system responds to an input. We study the response of these units
to a commonly occurring input - an edge. Specifically the inputs are
given by the following function.
\begin{eqnarray}
X(x,y) &=& e^{-0.25 v^2} \sin(v)\\
v &=& \vec{k} \cdot \vec{r} + k b \\
\vec{k} &=& k (\cos \theta, \sin \theta) \mbox{  and  } \vec{r} = (x,y)
\label{eq_edge}
\end{eqnarray}
where $(x,y)$ is the position of a pixel from the center of a patch,
$b$ a real number specifying distance of the edge from the center and
$\theta$ is the orientation of the edge from the $x$ axis. This is not
an edge function, but a function obtained on an edge after local mean
subtraction.

The responses of the simple units and the invariant units are shown in
the Figure. \ref{fig_responses}, see caption for description. As
expected the sparse coding units respond to edges in a narrow range of
positions and relatively narrow range of orientations. Invariant cells
on the other hand are able to pool different sparse coding units
together and become invariant to a larger range of positions. Thus the
invariant units do indeed have the desired invariance properties.

Note that for large sparsities the regions have clear boundaries and
are quite sharp. This is similar to the standard implementation of
convolutional net, where the pooling regions are squares (with clear
boundaries). It is probably more preferable to have regions that
overlap as happens at lower sparsities since one would prefer smoother
responses rather then jumps across boundaries.

\section{Theoretical analysis of the full model and its multi-layer generalization}
\label{sec_theoretical}

In this section we return to the full model (\ref{eq_fullE}). We
generalize it to an $n$ layer system, give an inference algorithm and
outline the proof that under certain assumptions of convexity the
algorithm has the fast $1/k^2$ convergence of FISTA, there $k$ is the
iteration number.

The basic idea of minimizing over $z,u$ in (\ref{eq_fullE}) is to
alternate between taking energy-lowering step in $z$ while fixing $u$
and taking energy-lowering step in $u$ while fixing $z$. Note that
both of the restricted problems (problem in $z$ fixing $u$ and problem
in $u$ fixing $z$) satisfy conditions of the FISTA algorithm. This
will allow us to take steps of appropriate size that are guaranteed to
lower the total energy. Before that however, we generalize the
problem, which will reveal its structure and which does not introduce
any additional complexity.

Consider system consisting of $n$ layers with units $z_a$ in the
$a$-$th$ layer with $a=1,\ldots,0$. We define $z=(z_1,\ldots,z_n)$
that is all the vectors $z_a$ concatenated. We define two sets of
functions. Let $e_a^{\beta}(z_a)$ be continuously differentiable,
convex and Lipschitz functions. There can be several such functions
per layer, which is denoted by index $\beta$. Let $g_a^{\beta}(z_a)$
be continuous and convex functions, not necessarily smooth. For
convenience we define $z_0=\emptyset$, $z_{n+1}=\emptyset$, $g_0=1$
and $e_{n+1}=1$.

We define the energy of the system to be
\begin{equation}
E = \sum_{a=0}^n \sum_{\beta} g_a^{\beta} (z_a) e_{a+1}^{\beta}
(z_{a+1}) \equiv \sum_{a=0}^n g_a (z_a) e_{a+1} (z_{a+1})
\label{eq_hierarchy}
\end{equation}
where in the second equality we drop the $\beta$ from the notation for
simplicity. We will omit writing the $\beta$ for the rest of the
paper. The equation (\ref{eq_fullE}) is a special case of
(\ref{eq_hierarchy}) with $n=2$, $g_0=1$,$e_1(z_1)=(1/2)\|x-Wz_1\|^2$,
$g_1(z_1)=|z_1|$, $e_2(z_2)=\alpha(1+ e^{-Az_2})/2$, $g_2(z_2)= \beta
|z_2|$ with $z_2 \geq 0$ and $e_3=1$.

Now observe that given $a \in \{1,\ldots,n\}$ the problem in $z_a$
keeping other variables fixed satisfies the conditions of the FISTA
algorithm. We can define a step in the variable $z_a$ to be (in
analogy to \cite{FISTA} eq. (2.6)):
\begin{eqnarray}
p_{L_a}(z_a) &=& \mbox{argmin}_{z'_a} \lbrace
g(z'_a)e_{a+1}(z_{a+1})+\frac{L}{2} \left|z'_a-(z_a-\frac{1}{L}
g_{a-1}(z_{a-1}) \nabla e_a(z_a)) \right|^2\rbrace \nonumber\\ &=&
\mbox{sh}_{\frac{e_{a+1}(z_{a+1})}{L_a}} \left(z_a-\frac{1}{L_a}
g_{a-1}(z_{a-1}) \nabla e_a (z_a) \right)
\label{eq_stepa}
\end{eqnarray}
where the later equality holds if $g_a(z_a)=|z_a|$. Here sh is the
shrinkage function
$\mbox{sh}_\alpha(z)=\mbox{sign}(z)(|z|-\alpha)_{+}$. In the case
where $z_a$ is restricted to be nonnegative we need to use
$(z-\alpha)_{+}$ instead of the shrinkage function.

Let us describe the algorithm for minimizing (\ref{eq_hierarchy}) with
respect to $z$ (we will write it explicitly below). In the order from
$a=1$ to $a=n$, take the step $z'_a=p_{L_a}(z_a)$ in
($\ref{eq_stepa}$). Repeat until desired accuracy. The $L_a$'s have to
be chosen so that
\begin{eqnarray}
&& g_{a-1}(z_{a-1})e_a(z'_a) \leq g_{a-1}(z_{a-1})e_a(z_a) + \nonumber \\
&& \langle z'_a-z_a, g_{a-1}(z_{a-1}) \nabla e_a(z_a) \rangle + (L/2) |z'_a-z_a|^2
\label{eq_pproperty}
\end{eqnarray}
This can be assured by taking $L_a \geq L(g_{a-1}e_{a})$ where the
later $L$ denotes the Lipschitz constant of its argument. Otherwise,
as used in our simulations, it can be found by backtracking, see
below. This will assure that each steps lowers the total energy
(\ref{eq_hierarchy}) and hence the overall procedure will keep
lowering it. In fact the step $p_L$ with such chosen $L$ is in some
sense a step with ideal step size. Let us now write the algorithm
explicitly:

{\bf Hierarchical (F)ISTA.}\\ Step 0. Take $L_0^a > 0$, some $\eta >
1$ and $\tilde{z}_0^a \in \Re$. Set $z_1 = \tilde {z}_0^a$, $t_1=1$,
$a=1,\ldots,n$. \\ Step k. ($k \geq 1$). \\ Loop a=1:n \{ Backtracking
\{ \\ .\hspace{6mm} Find smallest nonnegative integer $i_k^a$ such
that with $\bar{L}^a=\eta^{i_k^a} L^a_{k-1}$
\begin{eqnarray}
g_{a-1}(\tilde{z}_k^{a-1})e_a(p_{\bar{L}^a}(z_k^a))
&\leq& g_{a-1}(\tilde{z}_k^{a-1}) e(z_k^a) + \\
&& \langle p_{L^a}(z_k^a)-z_k^a, g_a(z_k^{a-1}) \nabla e(z_k^a) \rangle + \frac{1}{2} |p_{\bar{L}^a}(z_k^a)-z_k^a|^2
\end{eqnarray}
\hspace{6mm} Set $L_k^a = \eta^{i_k^a} L_{k-1}^a$ \}\\
.\hspace{3mm} Compute
\begin{equation}
\tilde{z}_k^a = p_{L_k^a}(z_k^a) \hspace{10mm} \}
\end{equation}
\begin{equation}
r_k=0 \hspace{5mm} \mbox{or} \hspace{5mm} t_{k+1} = \frac{1+\sqrt{1+4t_k^2}}{2}; \hspace{2mm} r_k=\frac{t_k-1}{t_{k+1}}
\label{eq_rk}
\end{equation}
\begin{equation}
\mbox{Loop a=1:n} \{ \hspace{5mm} z_{k+1}^a = \tilde{z}_k^a + r_k(\tilde{z}_k^a - \tilde{z}_{k-1}^a) \hspace{5mm} \}
\end{equation}

The algorithm described above is this algorithm with the choice
$r_k=0$ in the second last line.

Let us discuss the $r_k$'s. For single layer system ($n=1$) $r_k=0$
choice is called ISTA and has convergence bound of $E(z_k)-E(z^{*})
\leq \alpha L \|z_0-z^{*}\|^2/2k$. The other choice of $r_k$ is the
FISTA algorithm. The convergence rate of FISTA is much better than
that of ISTA, having $k^2$ in the denominator.

For hierarchical system, the choice $r_k=0$ guarantees that each step
lowers the energy. The question is whether introducing the other
choice of $r_k$ would speed up convergence to the FISTA convergence
rate. The trouble is that the in general the product $ge$ is
non-convex, which is the case for (\ref{eq_fullE}). For example we can
readily see that if the function has more then one local minima, this
convergence would certainly not be guaranteed (imagine starting at a
non-minimal point with zero derivative). The effect of $r_k$ is that
of momentum and this momentum increases towards one as $k$
increases. With such a large momentum the system is in a danger of
``running around''. It might be effective to introduce this momentum
but to regularize it (say bound it by a number smaller then one). In
any case one can always use the algorithm with $r_k=0$.

In the special cases when all $ge$'s are convex however, we give an
outline of the proof that the algorithm converges with the FISTA rate
of convergence. For this purpose we define the full step in $z$,
$P_L(z)$, to be the result of the sequence of steps
$z'_a=p_{L_a}(z_a)$ eq. (\ref{eq_stepa}) from $a=1$ to $a=n$. That is
we have $P_L$: $(z_1,z_2,\ldots,z_n) \rightarrow (z'_1,z_2,\ldots,z_n)
\rightarrow (z'_1,z'_2,\ldots,z_n) \rightarrow
(z'_1,z'_2,\ldots,z'_n)$. We assume that all the $L_a$'s are the same
(this is always possible by making all $L_a$'s equal the largest
value).

The core of the proof is to show the Lemma 2.3 of \cite{FISTA}:

{\bf Lemma 2.3:} Assume that $\forall a \in \{0,\ldots,n\}$, $e_a$ is
continuously differentiable, Lipschitz, $g_a$ is continuous, $g_a
e_{a+1}$ is convex and $P_L$ is defined by the sequence of $p_L$'s of
(\ref{eq_stepa}) as described above. Then for any $z,\hat{z}$
\begin{equation}
E(\hat{z})-E(P_L(z)) \geq \frac{L}{2} |P_L(z)-z|^2+L \langle z-\hat{z}, P_L(z)-z\rangle
\label{eq_lemma23}
\end{equation}

The proof in \cite{FISTA} shows that if the algorithm consists of
applying the sequence of $P_L$'s and these $P_L$'s satisfy Lemma 2.3,
then the algorithm converges with rate $E(z_k)-E(z_{*}) \leq
2L_{\mbox{max}} (z_0-z_{*})^2/(k+1)^2$. Thus we need to prove Lemma
2.3. We start with the analog of Lemma 2.2 of (\cite{FISTA}).

{\bf Lemma 2.2:} For any $z$, one has $z_a'=p_L^a(z_a)$ if and only if
there exist $\gamma_a (z_a) \in \partial g_a(z_a)$, the subdifferential of
$g_a(\cdot)$, such that
\begin{equation}
g(z_{a-1}) \nabla e(z_a) + L(z'_a-z_a)+ e(z_{a+1}) \gamma (z_a) = 0
\end{equation}
This lemma follows trivially from the definition of $p_{L_a}(z_a)$ as in \cite{FISTA}.

{\bf Proof of Lemma 2.3:} Define $z'=P_L(z)$. From convexity we have
\begin{eqnarray}
g_a(\hat{z}_a)e_{a+1}(\hat{z}_{a+1}) &\geq & g_a(z'_a)e_{a+1}(z_{a+1})
+ \langle \hat{z}_a - z'_a, e_{a+1}(z_{a+1}) \gamma_a (z_a) \rangle +
\nonumber \\ && \langle z'_{a+1} - z_{a+1}, g_a(z'_a) \nabla
e_{a+1}(z_{a+1}) \rangle
\label{eq_convex}
\end{eqnarray}
Next we have the property (\ref{eq_pproperty}). However the $z_{a-1}$
should be primed ($z'_{a-1}$) because the $z_{a-1}$ has already been
updated. Due to space limitations we won't write out all the
calculations but specify the sequence of operations. The details are
written out in the supplementary material. We take the first term on
the left side of (\ref{eq_lemma23}), $E(\hat{z})$ and express it in
its terms (\ref{eq_hierarchy}). Then, replace the terms using the
convexity equations and substitute $\gamma(z_a)$'s using the Lemma
2.2. Then we take the second term of the left side of
(\ref{eq_lemma23}), $E(P_L(z))$, again express it using
(\ref{eq_hierarchy}), and use the inequalities
(\ref{eq_pproperty}). Putting it all together, all the gradient terms
cancel and the other terms combine to give Lemma 2.3. This completes
the proof.

\section{Conclusions}
We introduced simple and efficient algorithm from learning invariant
representation from unlabelled data. The method takes advantage of
temporal consistency in sequential image data. In the future we plan
to use the invariant features discovered by the method to hierarchical
vision architectures, and apply it to recognition problems.


\small{
\bibliographystyle{unsrt}
\bibliography{sparse_invariance}
}

\newpage

\appendix

\section{Supplementary material for Efficient Learning of Sparse Invariant Representations}
1) We give details of the Proof of Lemma 2.3. \\
2) We show all of the invariant filters for system of: 20x20 patches input patches, 1000 simple units, 400 invariant units.

\subsection{Lemma 2.3}

{\bf Lemma 2.3:} Assume that $\forall a \in \{0,\ldots,n\}$, $e_a$ is continuously differentiable, Lipschitz, $g_a$ is continuous, $g_a e_{a+1}$ is convex and $P_L$ is defined by the sequence of $p_L$'s in the paper. Then for any $z,\hat{z}$
\begin{equation}
E(\hat{z})-E(P_L(z)) \geq \frac{L}{2} |P_L(z)-z|^2+L \langle z-\hat{z}, P_L(z)-z\rangle
\label{eq_lemma23}
\end{equation}

{\bf Proof of Lemma 2.3:}
Define $z'=P_L(z)$. We first collect the inequalities that we will need. \\
From convexity we have
\begin{eqnarray}
g_a(\hat{z}_a)e_{a+1}(\hat{z}_{a+1}) &\geq & g_a(z'_a)e_{a+1}(z_{a+1}) + \langle \hat{z}_a - z'_a, e_{a+1}(z_{a+1}) \gamma_a (z_a) \rangle + \nonumber \\
&& \langle z'_{a+1} - z_{a+1}, g_a(z'_a) \nabla e_{a+1}(z_{a+1}) \rangle
\label{eq_convex}
\end{eqnarray}
Next we have the property for step $p_{L_a}$ that guarantees that the energy is lowered in each step.
\begin{eqnarray}
&& g_{a-1}(z'_{a-1})e_a(z'_a) \leq g_{a-1}(z'_{a-1})e_a(z_a) + \nonumber \\
&& \langle z'_a-z_a, g_{a-1}(z'_{a-1}) \nabla e_a(z_a) \rangle + (L/2) |z'_a-z_a|^2
\label{eq_pproperty}
\end{eqnarray}
Finally we have the Lemma 2.2
\begin{equation}
g(z'_{a-1}) \nabla e(z_a) + L(z'_a-z_a)+ e(z_{a+1}) \gamma (z_a) = 0
\label{eq_gamma}
\end{equation}
Now we can put these equations together. The steps are: Write out the left side of (\ref{eq_lemma23}) in terms of the definition of E. Use inequalities (\ref{eq_convex}) and (\ref{eq_pproperty}). Eliminate $\gamma$'s using (\ref{eq_gamma}). Simplify. Here are the details:
\begin{eqnarray}
E(\hat{z})-E(z') &=& \sum_a g_a(\hat{z}_a)e_{a+1}(\hat{z}_{a+1}) - g_a(z'_a)e_{a+1}(z'_{a+1}) \nonumber \\
& \geq & \sum_a g_a(z'_a)e_{a+1}(z_{a+1}) + \langle \hat{z}_a - z'_a, e_{a+1}(z_{a+1}) \gamma_a (z_a) \rangle + \langle z'_{a+1} - z_{a+1}, g_a(z'_a) \nabla e_{a+1}(z_{a+1}) \rangle \nonumber \\
&& - g_{a-1}(z'_{a-1})e_a(z_a) -
 \langle z'_a-z_a, g_{a-1}(z'_{a-1}) \nabla e_a(z_a) \rangle - (L/2) |z'_a-z_a|^2 \nonumber \\
&=& \sum_a g_a(z'_a)e_{a+1}(z_{a+1}) - L \langle \hat{z}_a - z'_a, z'_a-z_a \rangle - \langle \hat{z}_a - z'_a,g(z'_{a-1}) \nabla e(z_a) \rangle \nonumber \\
&& + \langle z'_{a+1} - z_{a+1}, g_a(z'_a) \nabla e_{a+1}(z_{a+1}) \rangle \nonumber \\
&& - g_{a-1}(z'_{a-1})e_a(z_a) -
 \langle z'_a-z_a, g_{a-1}(z'_{a-1}) \nabla e_a(z_a) \rangle - (L/2) |z'_a-z_a|^2 \nonumber \\
&=& \sum_a - L \langle \hat{z}_a - z'_a, z'_a-z_a \rangle - (L/2) |z'_a-z_a|^2 \nonumber \\
&=& - L \langle \hat{z} - z', z'-z \rangle - (L/2) |z'-z|^2 \nonumber \\
&=& (L/2) |z'-z|^2 + L\langle z-\hat{z},z'-z \rangle
\end{eqnarray}
which is the formula (\ref{eq_pproperty}).  Note that in the line 5 and in the first term of lines 6 we shifted $a$ by one. This completes the proof.

\subsection{Simple unit and invariant unit filters. $\alpha=0.5$, $\beta=0.3$}

\begin{figure}[h]
\begin{center}
\centerline{\includegraphics[width=\columnwidth]{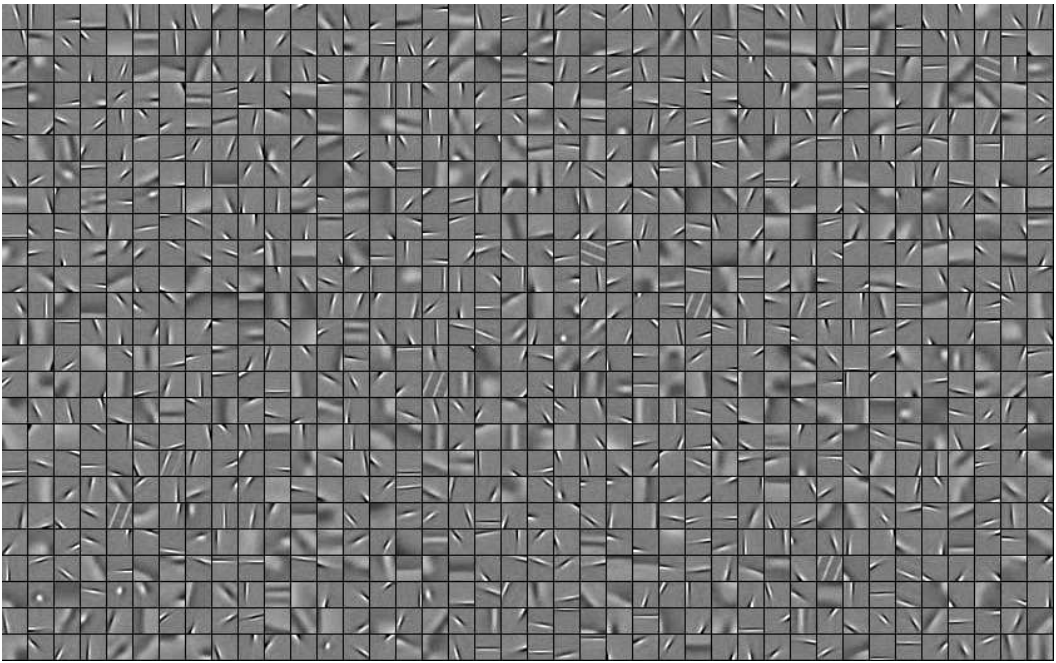}}
\end{center}
\caption{Sparse coding filters. Inputs 20x20 images patches, preprocessed. code: 1000 simple units.}
\label{fig_sc}
\end{figure}

\begin{figure}[h]
\begin{center}
\centerline{\includegraphics[width=\columnwidth]{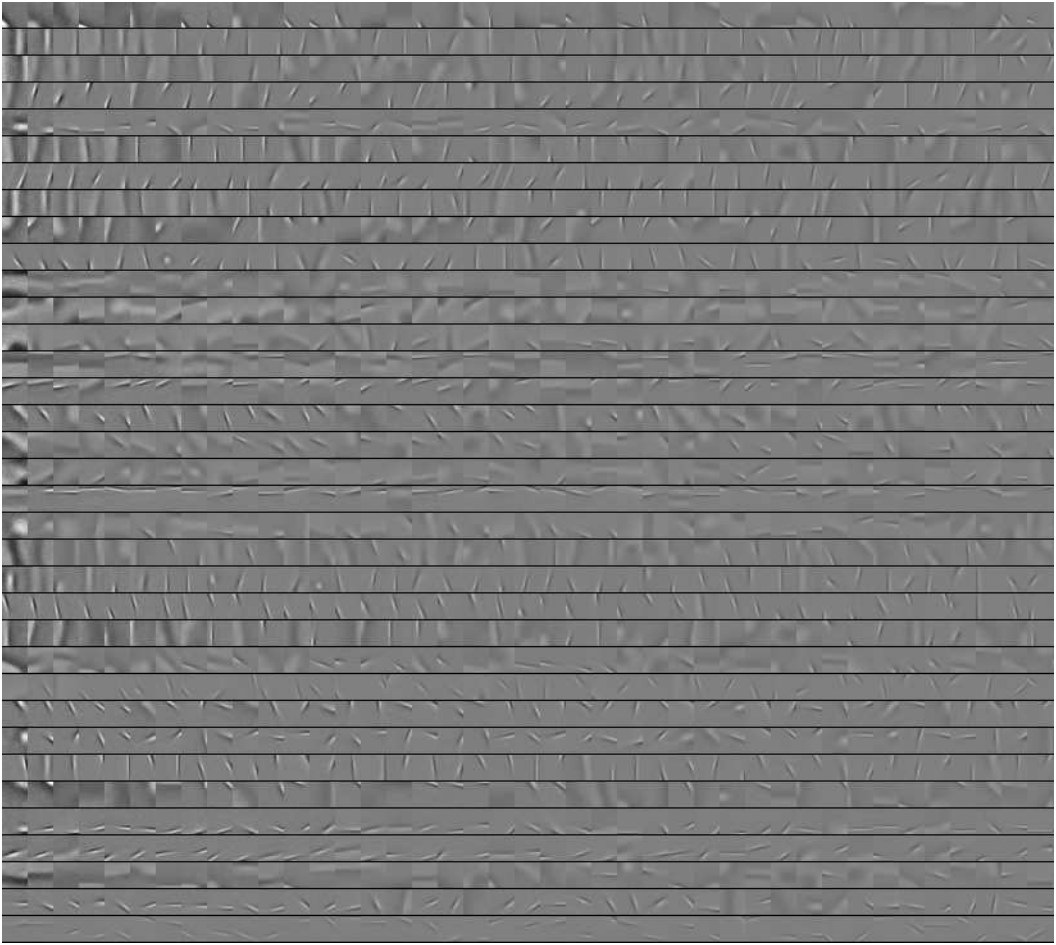}}
\end{center}
\caption{Selection of invariant units. See Figure 2a of the paper for explanation. System: 20x20 patches, 1000 simple units, 400 invariant units.}
\label{fig_ic}
\end{figure}

\begin{figure}[h]
\begin{center}
\centerline{\includegraphics[width=\columnwidth]{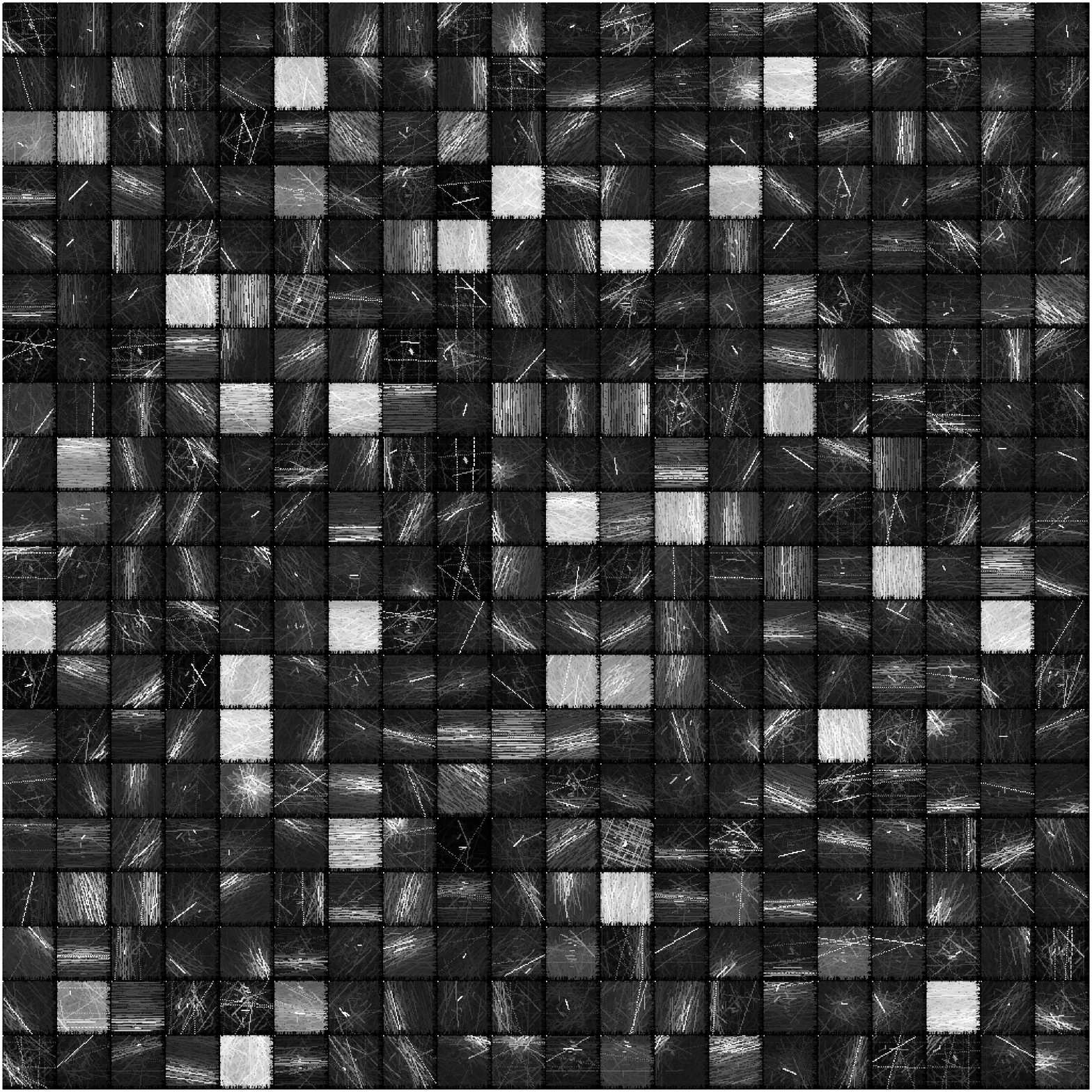}}
\end{center}
\caption{All 400 invariant units. See Figure 2b of the paper for explanation. System: 20x20 patches, 1000 simple units, 400 invariant units.}
\label{fig_lines}
\end{figure}

\begin{figure}[h]
\begin{center}
\centerline{\includegraphics[width=\columnwidth]{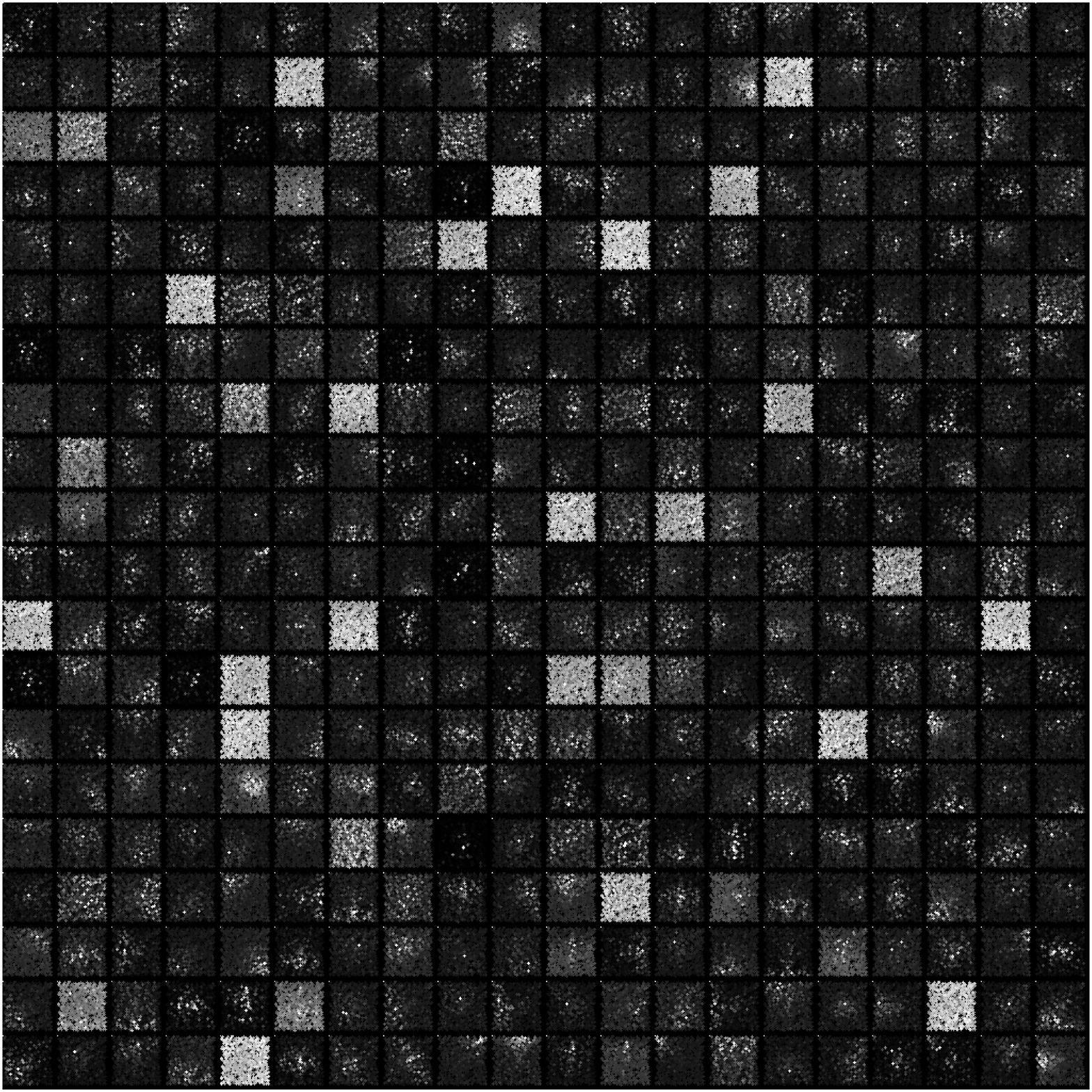}}
\end{center}
\caption{All 400 invariant units. Position of the invariant cells (centers of gabor fits). System: 20x20 patches, 1000 simple units, 400 invariant units.}
\label{fig_pos}
\end{figure}

\begin{figure}[h]
\begin{center}
\centerline{\includegraphics[width=\columnwidth]{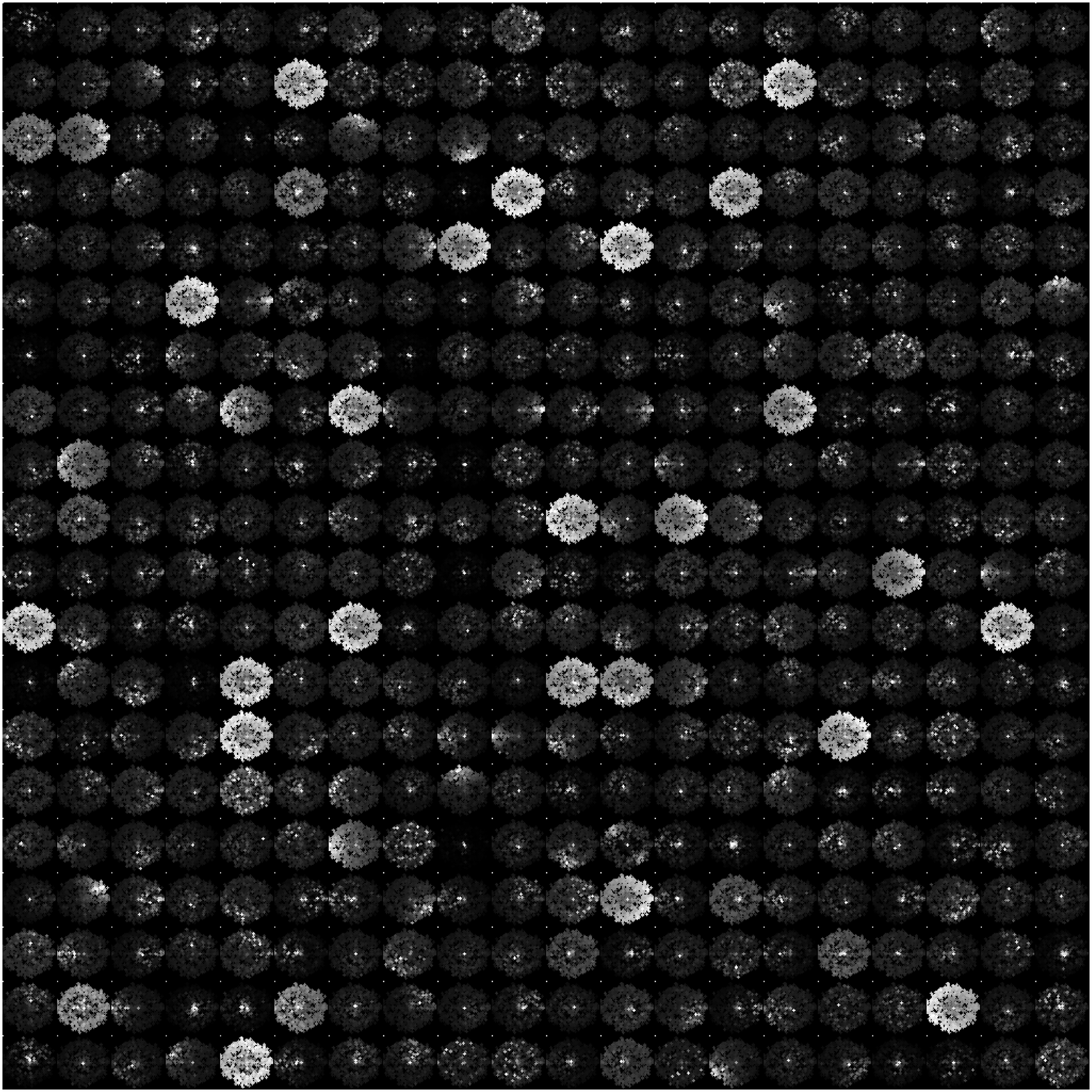}}
\end{center}
\caption{All 400 invariant units. See Figure 2c of the paper for explanation. System: 20x20 patches, 1000 simple units, 400 invariant units.}
\label{fig_ang_freq}
\end{figure}


\end{document}